\documentclass[runningheads]{llncs}

\usepackage{multirow}
\usepackage{epsfig}
\usepackage{amsmath}
\usepackage{amsfonts}
\usepackage{float}
\usepackage{booktabs}
\usepackage{color}
\usepackage[colorlinks]{hyperref}

\begin{document}
\title{DATR: Domain-adaptive transformer for \\multi-domain landmark detection}
% Anatomy-aware

\author{Heqin Zhu\inst{1} \and Qingsong Yao\inst{1} \and S.kevin Zhou\inst{1,2}}
\authorrunning{H. Zhu et al.}

%\email{\{zhuheqin20,yaoqingsong19\}@mails.ucas.edu.cn}\\
%\email{andrew.lxiao@gmail.com}\\
%\email{s.kevin.zhou@gmail.com}

\institute{\
Key Lab of Intelligent Information Processing of Chinese Academy of Sciences (CAS), Institute of Computing Technology, CAS, Beijing 100190, China\\
\email{s.kevin.zhou@gmail.com}  \and
School of Biomedical Engineering \& Suzhou Institute for Advanced Research,
Center for Medical Imaging, Robotics, and Analytic Computing \& LEarning (MIRACLE),
University of Science and Technology of China, Suzhou 215123, China}

\maketitle              % typeset the header of the contribution
\begin{abstract}    
Accurate anatomical landmark detection plays an increasingly vital role in medical image analysis. Although existing methods achieve satisfying performance, they are mostly based on CNN and specialized for a single domain say associated with a particular anatomical region. In this work, we propose a universal model for multi-domain landmark detection by taking advantage of transformer for modeling long dependencies and develop a domain-adaptive transformer model, named as DATR, which is trained on multiple mixed datasets from different anatomies and capable of detecting landmarks of any image from those anatomies. The proposed DATR exhibits three primary features: (i) It is the first universal model which introduces transformer as an encoder for multi-anatomy landmark detection; (ii) We design a domain-adaptive transformer for anatomy-aware landmark detection, which can be effectively extended to any other transformer network; (iii) Following previous studies, we employ a light-weighted guidance network, which encourages the transformer network to detect more accurate landmarks. We carry out experiments on three widely used X-ray datasets for landmark detection, which have 1,588 images and 62 landmarks in total, including three different anatomies (head, hand, and chest). Experimental results demonstrate that our proposed DATR achieves state-of-the-art performances by most metrics and behaves much better than any previous convolution-based models. The code will be released publicly.

\end{abstract}
\section{Introduction}
Anatomical landmark detection, aiming at locating key points in a medical image, contributes to various medical image analysis tasks~\cite{zhou2019handbook,zhou2021review}, such as ultrasound-probe movement guidance~\cite{usprobe}, segmentation~\cite{ref_seg} and registration~\cite{ref_reg}. To automatically and robustly detect accurate landmarks, a multitude of CNN-based methods have been developed, which are based on global-local scheme~\cite{ref_scn}, graph convolution~\cite{lang2020automatic}, multi-task learning~\cite{ref_mtn,qsyao2020landmarkattack}, uncertainty modeling~\cite{browning2021uncertainty}, few-shot learning~\cite{yao2021one}, etc. Although previous methods have achieved satisfying performances, the performance still can be improved from the following two aspects.

Firstly, previous methods are mostly specialized to a single task associated with a specific domain or anatomical region, unable to work on a new task or domain, and cannot benefit from the common knowledge among different domains or anatomies. With the development of multi-domain learning, Huang \textit{et al.}~\cite{ref_u2net} propose a universal U-Net based on a domain adaptor and achieves comparable performances in medical segmentation on multiple domains. Taking advantage of domain adaptor for multi-domain learning and dilated convolution for global information, Zhu \textit{et el.}~\cite{gu2net} propose the first universal model called GU2Net for multi-domain landmark detection, which outperforms landmark detection methods in single domain. Inspired by GU2Net, our proposed method consists of domain-specific and domain-shared parameters, aiming at learning domain-specific features and common knowledge, respectively. Moreover, following the idea of GU2Net, we train model from mixed datasets on different domains and leverages the common knowledge for improved performance.

Secondly, blooming approaches utilize transformer~\cite{Transformer} to model long-range dependencies and improve performances for various tasks in medical imaging analysis. 
Zhang \textit{et al.}~\cite{zhang2021multi} propose a multi-band hybrid transformer network for corneal endothelial cell segmentation. Feng \textit{et al.}~\cite{feng2021task} use transformer for joint MRI reconstruction and super-resolution, which embeds and synthesizes the relevance between the two tasks. Zhang \textit{et al.}~\cite{zhang2021learning} utilize the self-attention mechanism of transformer to model both the inter- and intra-image relevances for registration. Since accurate localization of landmarks deeply relies on global information~\cite{ref_scn,ref_mtn,gu2net}, it is promising to introduce the transformer for modeling long dependencies and extracting representative features. In this work, we leverage the advantages of transformer to encode the global context of landmarks for multi-domain landmark detection. 

In detail, our universal model, named as \textbf{D}omain-\textbf{A}daptive \textbf{TR}ansformer (\textbf{DATR}), is the first transformer-based method for multi-domain landmark detection, which consists of an transformer-based encoder along with a guidance network. The encoder is built up with domain-adaptive transformer blocks (DAT) to model the implicit relevance and global context of landmarks and further extract domain-adaptive features. The decoder is based on domain-adaptive convolutions (DAC) introduced in~\cite{gu2net,ref_u2net}. DAT and DAC consist of domain-adaptive parameters and domain-shared parameters which are learned on single specific domain and all domains respectively. The guidance network is composed of dilated convolutions~\cite{gu2net} to generate a guidance heatmap that contains the coarse but unambiguous landmarks.

{In summary, our contributions can be categorized as follows:
\begin{itemize}
    \item We propose the first domain-adaptive transformer network for multi-domain landmark detection, which effectively works well on multiple mixed datasets from various anatomical regions.
    
    \item The domain-adaptive transformer block (DAT) is effective for multi-domain learning and can be used in any other transformer network.
    
    \item Quantitative results demonstrate the effectiveness of our proposed DATR of detecting a total of 62 landmarks based on three publicly used X-ray datasets of head, hand, and chest.
    
    \item We carry out exhaustive experiments to demonstrate the effectiveness of learning on multiple domains and transferring to an unseen domain.
\end{itemize}
}

\section{Method}
\begin{figure}[t]
    \includegraphics[width=0.9\textwidth]{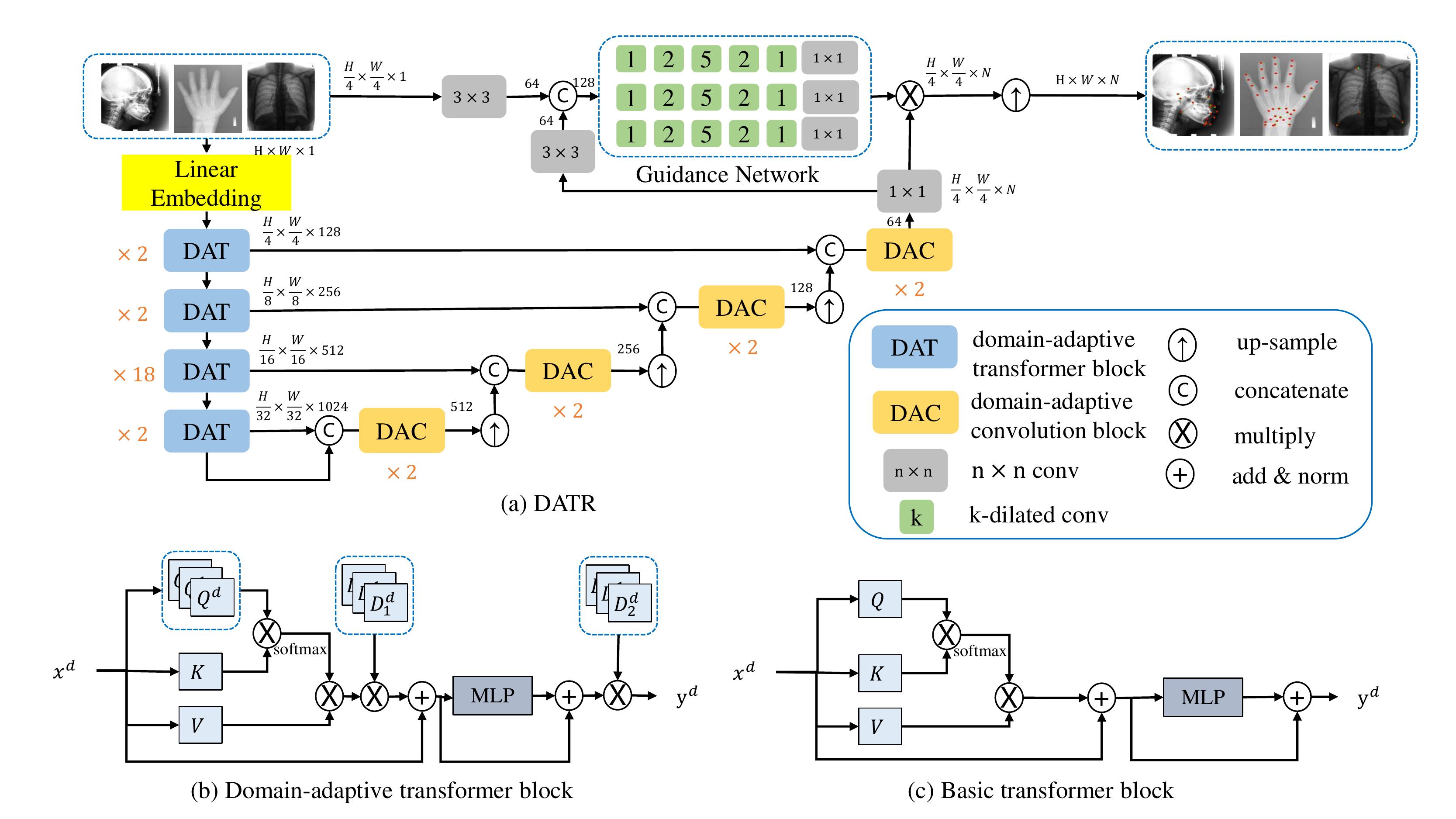}
    \caption{(a) The architecture of our proposed model, which is composed of domain-adaptive transformer encoder (DAT) and domain-adaptive convolution~\cite{ref_u2net,gu2net} decoder (DAC), additionally with guidance network~\cite{gu2net}. (b) Domain-adaptive transformer block. (c) Basic transformer block. Each domain-adaptive transformer is a basic transformer block with query matrix duplicated and domain-adaptive diagonal for each domain. The batch-normalization, activation and patch merging are omitted.}
    \label{fig_overall}
\end{figure}

Fig.~\ref{fig_overall} show that our proposed DATR model, of encoder-decoder architecture, takes advantage of both convolution and transformer for multi-domain landmark detection tasks. The encoder is composed of domain-adaptive transformer blocks based on Swin Transformer blocks~\cite{swin}, detailed in Section~\ref{sec:encoder}. The decoder is a U-Net~\cite{ref_unet} decoder with each $3\times 3$ convolution replaced with domain-adaptive convolution, which contains domain-specific channel-wise convolutions and a domain-shared point-wise convolution~\cite{ref_u2net,gu2net}. Following GU2Net~\cite{gu2net}, to guide DATR for more accurately detecting, we adopt a guidance network which consists of parallel sequences of five dilated convolutions and one $1\times 1$ convolution for each domain. The overall architecture is demonstrated in Section~\ref{sec:overall}.

\subsection{Domain-adaptive transformer encoder}\label{sec:encoder}
As Figure~\ref{fig_overall}(a) shows, the encoder is built up with domain-adaptive transformer blocks (DAT), making full use of the capability of transformer for modeling global dependencies and extracts multi-domain multi-scale representative features. 

\textbf{Domain-adaptive transformer block.}
As in Fig.~\ref{fig_overall}(c), a basic transformer block~\cite{Transformer} consists of a multi-head self-attention module (MSA), followed by a two-layer MLP with GELU activation. Furthermore, Layer Normalization (LN) is adopted before each MSA and MLP and a residual connection is adopted after each MSA and MLP. Given a feature map $x^d \in R^{h\times w\times c}$ from domain $d$ with height $h$, width $w$ and channels $c$, the output feature map of MSA $\hat{y}^d$ and MLP $y^d$ can be formulated as:
\begin{equation}
    \begin{split}
        \hat{y}^d &=  \text{MSA}(\text{LN}(x^d))+x^d\\
        %\text{softmax}(\frac{Qx^d K^Tx^d})Vx^d\\
        y^d &= \text{MLP}(\text{LN}(\hat{y}^d)) + \hat{y}^d 
    \end{split}
    \label{Eq:trans}
\end{equation}
where $\text{MSA} = \text{softmax}(QK^T)V$.

Based on Eq.~(\ref{Eq:trans}), we propose a novel domain-adaptive transformer block (DAT) for multi-domain learning. Similar to U2Net~\cite{ref_u2net} and GU2Net~\cite{gu2net}, we adopt domain-specific and domain-shared parameters in DAT. Since the attention probability is depended on query and key matrix which are symmetrical, we duplicate the query matrix for each domain to learn domain-specific query features and keep key and value matrix domain-shared to learn common knowledge and reduce parameters. Inspired by LayerScale~\cite{layerscale}, we further adopt learnable diagonal matrix~\cite{layerscale} after each MSA and MLP module to facilitate the learning of domain-specific features, which costs few parameters ($O(N)$ for $N\times N$ diagonal). Different from LayerScale~\cite{layerscale}, our domain-adaptive matrices (DAM) $D_1^d$ and $D_2^d$ are applied for each domain with $D_2^d$ applied after residual connection for generating more representative and direct domain-specific features. The above process can be formulated as:
\begin{equation}
    \begin{split}
        \hat{y}^d &=  D_1^d \times \text{MSA}_{Q^d}(\text{LN}(x^d))+x^d\\
        %\text{softmax}(\frac{Qx^d K^Tx^d})Vx^d\\
        y^d &= D_2^d \times(\text{MLP}(\text{LN}(\hat{y}^d)) + \hat{y}^d )
    \end{split}
\end{equation}
where $\text{MSA}_{Q^d} = \text{softmax}(Q^dK^T)V$.

\textbf{Multi-scale feature extraction.}
Similar to Swin Transformer~\cite{swin}, we adopt shifted window and limit self-attention within non-overlapping local windows for computation efficiency and the effectiveness of modeling. The encoder consists of four stages with the number of DAT being $2,2,18,2$ respectively. Firstly, the input image in shape of $H\times W\times 1$ from a random batch is partitioned into non-overlapping patches and linearly embedded into a shape of $\frac{H}{4}\times \frac{W}{4}\times 128$. Next, these patches are fed into cascaded transformer blocks at each stage, which are merged except the last stage. As a result of patch merging, the number of patches decreases by four times while channel dimension doubles. Finally, four scales of feature maps are generated with shapes being $\frac{H}{4}\times \frac{W}{4}\times 128$, $\frac{H}{8}\times \frac{W}{8}\times 256$, $\frac{H}{16}\times \frac{W}{16}\times 512$, and $\frac{H}{32}\times \frac{W}{32}\times 1024$, respectively.

%To work well on all datasets, we replace the standard convolution in U-Net with separable convolution which consists of domain-specific channel-wise convolution and shared point-wise convolution. Each data set is assigned a different channel-wise convolution separately, while all datasets share the same point-wise convolution.

\subsection{Overall pipeline} \label{sec:overall}
Given that a random input $X^d\in R^{H^d\times W^d \times C^d}$ belongs to domain $d$ from mixed datasets on various anatomical regions, which contains $N^d$ landmarks with corresponding coordinates being $\{(i_1^d,j_1^d), (i_2^d,j_2^d), \ldots, (i_{N_d}^d,j_{N_d}^d)\}$, we apply Gaussian function to obtain the $n^{th}$ ground truth heatmap $Y_{n}^d \in R^{H^d\times W^d \times C^d }$ as follows:
\begin{equation}
    Y_{n}^d= \frac{1}{\sqrt{2\pi}\sigma}\exp\{{-\frac{(i-i_{n}^d)^2+(j-j_{n}^d)^2}{2\sigma^2}}\},
\label{Eq:gt}
\end{equation}
where $n\in \{1,2,\dots, N^d\}$ and $\sigma$ is the standard deviation of Gaussian function.

As demonstrated in Figure~\ref{fig_overall}(a), the guidance network takes the down-sampled input in a shape of $\frac{H^d}{4}\times \frac{W^d}{4} \times C^d$ as input and generates a guidance heatmap $\tilde{Y}_{g}$. Meanwhile, with the original image as input, DATR employs domain-adaptive transformer encoder to extract four scales of distinguish features and passes them to domain-adaptive convolution decoder which concatenates in a cascade fashion, upsamples, converts multi-scale features, and generates a fine heatmap $\tilde{Y}_{f}$. The output accurate heatmap is produced by the pixel-multiplication of guidance heatmap and fine heatmap: $\tilde{Y} = \tilde{Y}_{g} \odot \tilde{Y}_{f}$. Finally, the landmark coordinates are extracted after finding the maximum location of the heatmap $\tilde{Y}$.

%where $\theta_{di}^L$ is the parameter of domain-specific channel-wise convolution corresponding to $D_i$; and $\theta_s^L$ is the parameter of shared point-wise convolution. In separable convolution, considering a N-channel input feature map and a M-channel output feature map,  we firstly apply $N$ channel-wise filters in the shape of $R^{3 \times 3}$ to each channel and concatenate the $N$ output feature maps. Secondly, we apply $M$ point-wise filters in shape of $R^{1\times 1\times N}$ to output the feature maps of $M$ channels~\cite{ref_u2net}. Accordingly, the total number of parameters is $9\times N\times T+N\times M$, while it's $9\times N\times M\times T$ for $T$ standard $3\times 3$ convolutions.

\section{Experiments}
\subsection{Setup}

\textbf{Datasets.} For performance evaluation, We adopt three widely used X-ray datasets from different domains on various anatomical regions of head, hand, and chest. (i) \underline{Head dataset}~\cite{ref_head} contains 400 X-ray cephalometric images with 150 images for training and 250 images for testing. Each image is of size $2400\times 1935$ with a resolution of $0.1mm \times 0.1mm$, which contains 19 landmarks manually labeled by two medical experts and we use the average labels same as Payer et al.~\cite{ref_scn}. (ii) \underline{Hand dataset}\footnote{\href{https://ipilab.usc.edu/research/baaweb}{https://ipilab.usc.edu/research/baaweb}} contains 909 X-ray images with 609 images for training and the other 300 images for testing. Each image contains 37 landmarks~\cite{ref_scn}. Following Payer et al.~\cite{ref_scn}, we calculate the physical distance as $\text{distance}_{\text{physical}}=\text{distance}_{\text{pixel}} \times \frac{50}{\|p-q\|_2}$  where $p,q$ are the two endpoints of the wrist respectively. (iii) \underline{Chest dataset}~\cite{gu2net} contains 279 X-ray images and is partitioned as 229 images for training and 50 images for testing, along with 6 landmarks in each image. Since physical spacing of chest dataset is not known, we use pixel distance for this dataset.

\noindent \textbf{Implementation details.}
Our model is implemented in Pytorch and trained on a TITAN RTX GPU with CUDA version being 11. Each $3\times 3 $ convolution is followed by batch normalization~\cite{ref_batnorm} and ReLU activation~\cite{ref_relu}. We initialize the encoder with pre-trained Swin Transformer. We resize each image to a shape of $512\times 512$ as the input of the model. At inference stage, the predicted heatmap is resized back to the origin shape for metric calculation. We set batch-size to 8 and set $sigma$ to 3. We adopt binary cross-entropy (BCE) loss and Adam optimizer to train the model for 100 epochs, with a cyclic scheduler~\cite{ref_clr} cyclicly adjust learning rate from 1e-4 to 5e-3. For evaluation, we choose model with minimum validation loss as the inference model and adopt two metrics: mean radial error (MRE) and successful detection rates (SDR). As deﬁned by Glocker et al.~\cite{glocker}, a predicted landmark is correctly identiﬁed if the MRE is small than 2mm, we follow it and set a threshold of 2mm for head, hand and 20px for chest. 

\begin{table}[t]
\centering
\caption{Quantitative comparison of our model with SOTA methods on head, hand, and chest datasets. \dag denotes the model is trained on single dataset respectively while \ddag denotes the model is learned on mixed datasets. - represents that no experimental results can be found in the original paper. In each column, the best results are in \textbf{bold} and the second best are \underline{underlined}. }
\label{tab_results}
\resizebox{1\linewidth}{!}{
\begin{tabular}{lrrrrrrrrrrrrr}
\hline
\multirow{2}*{Models}  &\multirow{2}*{MRE} &\multicolumn{4}{c}{Head  SDR(\%)}   &\multirow{2}*{MRE} & \multicolumn{3}{c}{Hand SDR(\%)} & \multirow{2}*{MRE} &  \multicolumn{3}{c}{Chest SDR(\%)}\\
\cline{3-6}\cline{8-10}\cline{8-10}\cline{12-14}   &(mm)& 2mm   & 2.5mm & 3mm   & 4mm   &(mm)     & 2mm & 4mm & 10mm  &(px)  & 3px & 6px & 9px\\
\hline
\multirow{1}*{Ibragimov et al.~\cite{ref_ibragimov}\dag}  &1.84& 68.13 & 74.63 & 79.77 & 86.87  &-    &-&-&-                  &-&-&-&-\\
%\multirow{1}*{{\v{S}}tern et al.~\cite{ref_stern}\dag}    &-&-&-&-&-                         &\underline{0.80} & 92.20 & 98.45 & 99.83 &-&-&-&-\\
\multirow{1}*{Lindner et al.~\cite{ref_lindner}\dag}      &1.67& 70.65 & 76.93 & 82.17 & 89.85 &0.85 & 93.68 & 98.95 & 99.94 &-&-&-&-\\
\multirow{1}*{Urschler et al.~\cite{ref_urschler}\dag}    &-& 70.21 & 76.95 & 82.08 & 89.01     &\underline{0.80}  & 92.19 & 98.46 & {99.95} &-&-&-&-\\
\multirow{1}*{Payer et al.~\cite{ref_scn}\dag}            &-& 73.33 & 78.76 & 83.24 & 89.75   &\textbf{0.66} & \underline{94.99} & \underline{99.27} & \textbf{99.99} &-&-&-&-\\
\multirow{1}*{U-Net~\cite{ref_unet}\ddag}                  &12.45& 52.08 & 60.04 & 66.54 & 73.68    &6.14  & 81.16 & 92.46 & 93.76 &{5.61} &\underline{51.67} & \underline{82.33} & \textbf{90.67}\\ 
\multirow{1}*{GU2Net~\cite{gu2net} \ddag}                        &\underline{1.54}& \underline{77.79} & \underline{84.65} & \underline{89.41} & \underline{94.93}   &0.84& \textbf{95.40} & \textbf{99.35} & {99.75} &\underline{5.57}& \textbf{57.33} &\textbf{82.67} & {89.33}\\
\hline

% run 0216
\multirow{1}*{DATR (Ours)\ddag}                        &\textbf{1.47}& \textbf{78.07} & \textbf{85.49} & \textbf{90.21} & \textbf{95.60}   &0.86& 94.04 & {99.20} & \underline{99.97} &\textbf{4.30}& {43.21} &{77.37} & \underline{90.53}\\
%id rate: head: 99.96   hand: 99.76   chest: 98.78
% std     head: 1.24    hand: 0.74    chest: 3.31

\hline
\end{tabular}

}
\end{table}

\subsection{Comparisons with state-of-the-art methods}
As Table~\ref{tab_results} shows, we involves 5 specilized models~\cite{ref_ibragimov,ref_stern,ref_lindner,ref_urschler,ref_scn} trained on single dataset separately and two models which employ multi-domain learning on mixed datasets (i.e., UNet~\cite{ref_unet}, GU2Net~\cite{gu2net}) for performance comparison. Our proposed DATR model achieves SOTA accuracy on head and comparable performances on hand and chest, outperforming any other convolution-based models whatever learned on a single dataset or mixed multiple datasets.

On the \underline{head dataset}, our model identifies correct landmarks with a rate of 99.96\%, obtaining an MRE of $1.47\pm 1.24$ mm, achieving the best accuracy within all four thresholds (i.e., 2mm, 2.5mm, 3mm, 4mm). Our model and GU2Net designed with domain-specific parameters and domain-shared parameters, employ multi-domain learning, resulting in much better performances than models trained on a single dataset. By comparing our model with GU2Net, our elaborately designed domain-adaptive transformer performs better than GU2Net which consists of convolutions. 
On the \underline{hand dataset}, our model identifies correct landmarks with a rate of 99.76\%, obtaining an MRE of $0.86\pm 0.74$mm. It's conceivable that our DATR performs slightly worse than the previous SOTA method GU2Net and SCN, since the hand dataset has a larger amount of samples in contrast to head, chest and benefit little from multi-domain learning. However, our DATR behaves much better than the other models.
On the \underline{chest dataset}, our model identifies correct landmarks with a rate of 98.78\%,  obtaining an MRE of $4.30\pm 3.31$ px, behaving better than other models on MRE by a huge gap. DATR obtains comparable SDR within 9px of 90.53\%, but a little worse on SDR within 3px and 6px than other models. It's probably a result of the big standard deviation since it is hard for Transformer network to train thoroughly on a limited scale of datasets.

\subsection{Ablation study}
\textbf{Effectiveness of DATR and guidance network.} The experiments are carried out on the mixed datasets which contain $250+300+50=600$ test images and 62 landmarks. We resize each image to $512\times 512$ and calculate the average MRE and SDR of all landmarks' pixel distance for measurement. The results are presented in Table~\ref{tab_ablation}.

\begin{table}[t]
\centering \scriptsize
\caption{Ablation study of different components of our DATR. $\text{GN}$ denotes the guidance network; $\text{MSA}_{Q^d}$ denotes the domain-adaptive self-attention and $D^d$ denotes the domain-adaptive diagonal matrix. In each column, the best results are in \textbf{bold}.}
\label{tab_ablation}
%\resizebox{1\linewidth}{!}
{
\begin{tabular}{c l ccc ccc ccc}
\hline
\multirow{2}*{Index} &\multirow{2}*{Models}& \multirow{2}*{MRE}  &\multicolumn{6}{c}{SDR(\%)}\\
\cline{4-9} & &px  &2px& 2.5px &3px  & 4px & 6px & 10px\\
\hline
(a)&Baseline                      &2.62 &46.40 &65.01 &71.07 &82.31 &92.53 &98.17 \\
(b)& +$\text{GN}$                 &2.61 &47.25 &65.26 &71.17 &82.28 &92.62 &98.08\\
(c)& +$\text{MSA}_{Q^d}$          &2.58 &47.22 &65.55 &71.33 &82.49 &92.72 &98.28 \\
(d)& +$D^d$                       &2.56 &48.28 &66.01 &71.65 &82.49 &92.53 &98.17\\
(e)& +$\text{MSA}_{Q^d}$+$D^d$    &2.53 &49.13 &66.54 &72.01 &82.85 &92.94 &98.26\\ 
(f)& +$\text{MSA}_{Q^d}$+$D^d$+$\text{GN}$ &\textbf{2.48} &\textbf{50.32} &\textbf{67.84} &\textbf{73.32} &\textbf{83.83} &\textbf{93.15} &\textbf{98.38}\\

\hline
\end{tabular}

}
\end{table}

\underline{Guidance network}. As shown in Table~\ref{tab_ablation}, Model (a), using basic transformer block and domain-adaptive convolution block as the component of encoder and decoder respectively, is adopted as the baseline. After using $\text{GN}$ to extract the guidance heatmap and guide the Baseline model, Model (b) performs a little better on all metrics, which demonstrates that global information is not only helpful for convolution networks~\cite{ref_scn,gu2net}, but also for transformer networks.

\underline{Domain-adaptive transformer}. The domain-adaptive transformer has two key improvements: domain-adaptive self-attention ($\text{MSA}_{Q^d}$) and domain-adaptive diagonal matrix ($D^d$). The performances of Model (c) and Model (d) which are much superior to that of Model (a) shows the effectiveness of $\text{MSA}_{Q^d}$ and $D^d$. Further, Model (e) combines the two and achieves much better performances. Compared to Model (a) using basic transformer block which obtains an MRE of 2.62px, Model (e), using domain-adaptive transformer block, obtains an MRE of 2.53px, which demonstrates the effectiveness of our proposed DATR.

\begin{figure*}[!t]
    \centering 
        \begin{minipage}[t]{0.12\textwidth}
        \centering
        \includegraphics[width=1\textwidth]{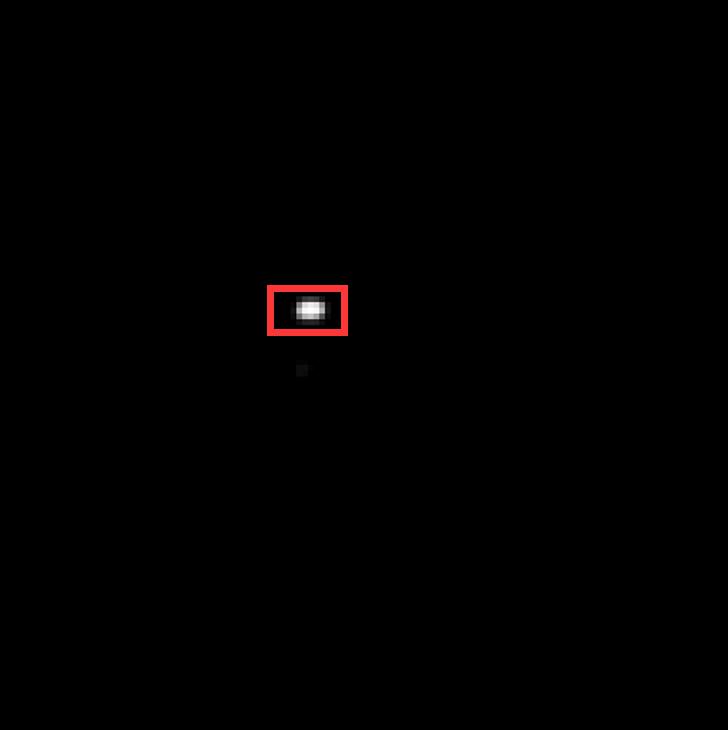}
        $\tilde{Y}_g$
    \end{minipage}
        \begin{minipage}[t]{0.12\textwidth}
        \centering
        \includegraphics[width=1\textwidth]{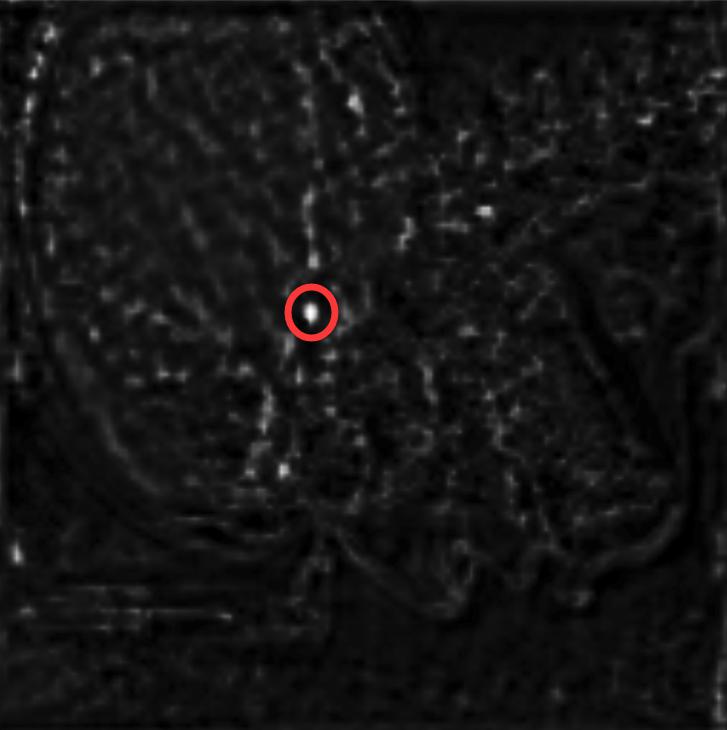}
        $\tilde{Y}_f$
    \end{minipage}
        \begin{minipage}[t]{0.12\textwidth}
        \centering
        \includegraphics[width=1\textwidth]{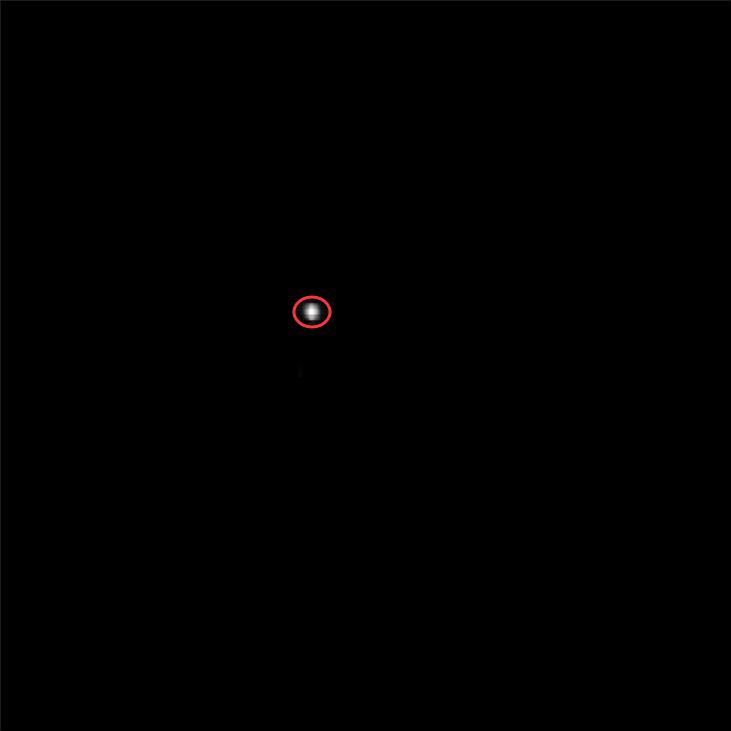}
        $\tilde{Y}$
    \end{minipage}
    \begin{minipage}[t]{0.12\textwidth}
        \centering
        \includegraphics[width=1\textwidth]{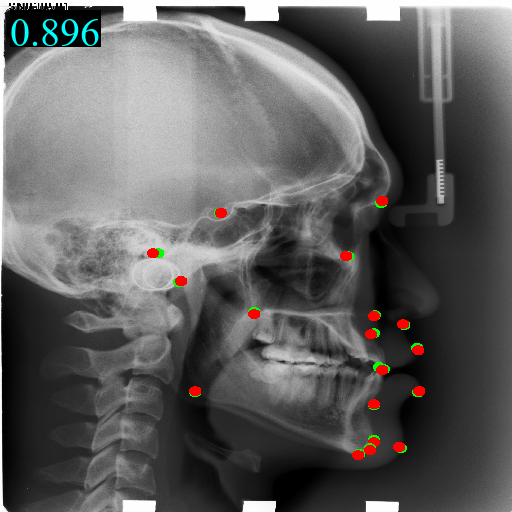}
        Head
    \end{minipage}
    \begin{minipage}[t]{0.12\textwidth}
        \centering
        \includegraphics[width=1\textwidth]{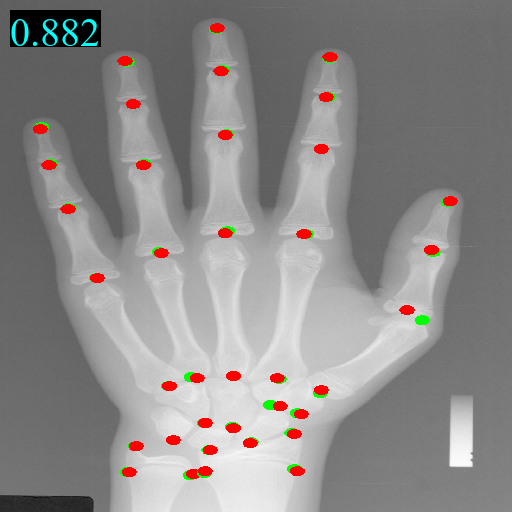}
        Hand
    \end{minipage}
    \begin{minipage}[t]{0.12\textwidth}
        \centering
        \includegraphics[width=1\textwidth]{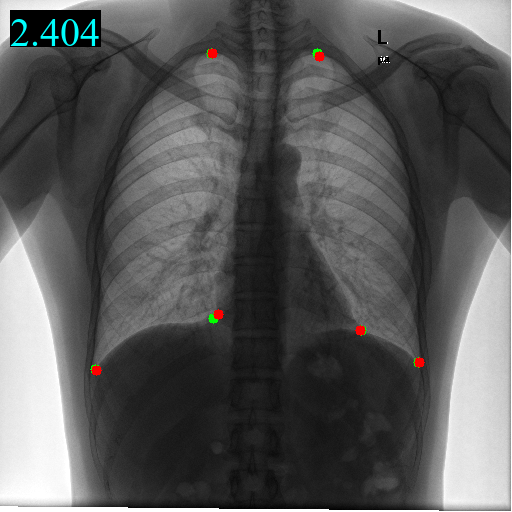}
        Chest
    \end{minipage}
        \begin{minipage}[t]{0.12\textwidth}
        \centering
        \includegraphics[width=1\textwidth]{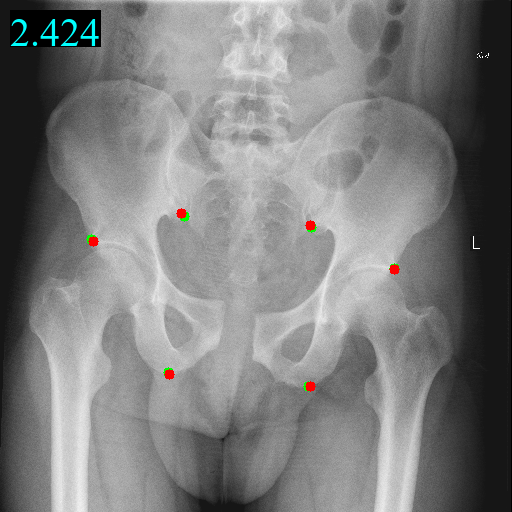}
        Pelvis
    \end{minipage}
    \caption{Visualization examples of guidance heatmap $\tilde{Y}_g$, fine heatmap $\tilde{Y}_f$, output heatmap $\tilde{Y}$ and images along with landmarks from four domains on head, hand, chest, and pelvis. The red points \textcolor{red}{$\bullet$} are predicted landmarks while the green points \textcolor{green}{$\bullet$} are annotated landmarks. The MRE value is on the top left corner of the image.}
    \label{fig_visual}
\end{figure*}

We take Model (f) which combines the above components as the final model. Model (f) using domain-adaptive transformer block and guidance network, beats any other models on all metrics by a huge gap. The predicted landmarks and heatmaps are exemplified in Figure~\ref{fig_visual}.

\noindent \textbf{Effectiveness of multi-domain learning}
To show the effectiveness of multi-domain learning and explore the learned domain-specific features and domain-shared features, we conduct experiments on different mixed datasets by combining a varying number of domains. We train our proposed DATR on: (1) a single dataset (head, hand, chest), (2) different combinations of two datasets (head+hand, head+chest, hand+chest), (3) three datasets (head+hand+chest), and test it on trained single dataset respectively. As Table~\ref{tab_domainwise} shows, (1) with the increasing number of training domains, MRE and SDR metrics generally become better on each dataset; (2) the model trained on all 3 datasets achieves the best performances in all metrics except SDR (4mm) on hand and SDR (3px) on chest; (3) the model trained on mixed datasets behaves generally better than model trained on single dataset. These results demonstrate learning common knowledge from multiple domains improves the detection accuracy of all datasets.

\begin{table}[t]
\centering
\caption{Evaluation of our proposed DATR trained on different number of mixed datasets and tested on single dataset respectively. In each column, the best results are in \textbf{bold} and the second best results are \underline{underlined}.}
\label{tab_domainwise}

\resizebox{1\linewidth}{!}
{
\begin{tabular}{l cccc cccc cccc}
\hline
\multirow{2}*{Datasets} &\multicolumn{4}{c}{Head}&\multicolumn{4}{c}{Hand} &\multicolumn{4}{c}{Chest} \\
\cline{2-13}      & MRE  & 2mm & 3mm & 4mm   & MRE  & 2mm & 4mm & 10mm   & MRE  & 3px & 6px & 9px\\
\hline
Head/Hand/Chest  & 1.60 & 74.17 & 87.71 & 94.45  & 0.90 & 93.20 & 99.18 & 99.76 & 4.51 & \underline{43.62} & 75.72 & 87.24\\
Head+Hand        & \underline{1.52} & \underline{76.94} &88.80 & 94.95 & \textbf{0.86} & \underline{93.69} & \textbf{99.26} & \textbf{99.97} & - & - &- &-      \\
Head+Chest       & 1.54 & 75.93 & \underline{89.80} & \underline{95.07} & - &- &-   & -     & 4.45 & 42.89 & 73.28 & 87.53\\
Hand+Chest       & -  & - & -  & -     & 0.87 & 93.47 & 99.08 & 99.95 & \underline{4.35} & \textbf{44.63} & \underline{75.82} & \underline{89.26}\\
Head+Hand+Chest  & \textbf{1.47} & \textbf{78.07} & \textbf{90.21} & \textbf{95.60} & \textbf{0.86} & \textbf{94.04}& \underline{99.20} & \textbf{99.97}  & \textbf{4.30} & 43.21 &\textbf{77.37} & \textbf{90.53}\\
\hline
\end{tabular}
}
\end{table}

\noindent \textbf{Transferring to a novel domain}
Furthermore, to demonstrate the effectiveness of new domain adaption, we finetune the trained DATR and corresponding basic transformer model on a new in-house pelvis dataset by freezing domain-shared parameters and adding domain-specific parameters in parallel. As Table ~\ref{tab_adaption} shows, our domain-specific transformer performs better than basic transformer, which indicates our proposed DATR learns shared common knowledge to boost the performance of each domain. The landmarks are visualized in Figure~\ref{fig_visual}.

\begin{table}[!t]
\centering \scriptsize
\caption{Quantitative results of domain adaption on a new pelvis dataset, including 6 landmarks, 100 training images and 36 test images.}
\label{tab_adaption}
%\resizebox{1\linewidth}{!}
{
\begin{tabular}{l c ccc}
\hline
Model& MRE (px)  &SDR\textless 3px(\%) &SDR \textless 6px(\%) &SDR \textless 9px(\%) \\
\hline
Basic transformer &5.64 &41.97 & 76.17 & 85.82 \\
Domain-adaptive transformer &\textbf{4.12} &\textbf{48.30} &\textbf{81.82} & \textbf{90.91}\\

\hline
\end{tabular}
}
\end{table}

\section{Conclusions}
To build a universal transformer model, we design a domain-adaptive transformer that extracts domain-specific features to facilitate the detection of landmarks, which deeply relies on the global context information. Our proposed model is the first transformer network in multi-domain landmark detection and the proposed domain-adaptive transformer block works with any other transformer network and domain-adaptive tasks. Experimental results show the effectiveness of our proposed model qualitatively and quantitatively, which achieves SOTA performances and beats previous convolution-based models by a big gap. Future work includes better design of transformer-based adaptive mechanism.

\bibliographystyle{splncs04}
\bibliography{paper}

\end{document}